\documentclass[final]{cvpr}

\usepackage{times}
\usepackage{epsfig}
\usepackage{graphicx}
\usepackage{amsmath}
\usepackage{amssymb}

\usepackage{enumitem}
\usepackage{multirow}
\usepackage{color}
\usepackage{pifont}

\usepackage{float}
\newfloat{figtab}{htb}{fgtb}
\makeatletter
\newcommand\figcaption{\def\@captype{figure}\caption}
\newcommand\tabcaption{\def\@captype{table}\caption}
\makeatother

\usepackage[pagebackref=true,breaklinks=true,colorlinks,bookmarks=false]{hyperref}

\def\cvprPaperID{7833} 
\def\confYear{CVPR 2021}

\begin{document}

\title{Towards Improving the Consistency, Efficiency, and Flexibility of \\ Differentiable Neural Architecture Search\vspace{-1mm}}

\author{
	Yibo Yang\textsuperscript{\rm 1,2},
	Shan You\textsuperscript{\rm 3},
	Hongyang Li\textsuperscript{\rm 2},
	Fei Wang\textsuperscript{\rm 3}, 
	Chen Qian\textsuperscript{\rm 3}, 
	Zhouchen Lin\textsuperscript{\rm 2,}\\ 
	\textsuperscript{\rm 1}Center for Data Science, Academy for Advanced Interdisciplinary Studies, Peking University\\
	\textsuperscript{\rm 2}Key Laboratory of Machine Perception (MOE), School of EECS, Peking University\\
	\textsuperscript{\rm 3}SenseTime\\
	{\{ibo, lhy\_ustb, zlin\}@pku.edu.cn}, {\{youshan, wangfei, qianchen\}@sensetime.com} 
	\vspace{-1mm}
}

\maketitle

\begin{abstract}
\vspace{-1mm}
Most differentiable neural architecture search methods construct a super-net for search and derive a target-net as its sub-graph for evaluation. There exists a significant gap between the architectures in search and evaluation. As a result, current methods suffer from an inconsistent, inefficient, and inflexible search process. In this paper, we introduce EnTranNAS that is composed of \underline{\textbf{En}}gine-cells and \underline{\textbf{Tran}}sit-cells. The Engine-cell is differentiable for architecture search, while the Transit-cell only transits a sub-graph by architecture derivation. Consequently, the gap between the architectures in search and evaluation is significantly reduced. Our method also spares much memory and computation cost, which speeds up the search process. A feature sharing strategy is introduced for more balanced optimization and more efficient search. Furthermore, we develop an architecture derivation method to replace the traditional one that is based on a hand-crafted rule. Our method enables differentiable sparsification, and keeps the derived architecture equivalent to that of Engine-cell, which further improves the consistency between search and evaluation. Besides, it supports the search for topology where a node can be connected to prior nodes with any number of connections, so that the searched architectures could be more flexible. For experiments on CIFAR-10, our search on the standard space requires only 0.06 GPU-day. We further have an error rate of 2.22\% with 0.07 GPU-day for the search on an extended space. We can also directly perform the search on ImageNet with topology learnable and achieve a top-1 error rate of 23.8\% in 2.1 GPU-day.
\end{abstract}

\vspace{-4mm}
\section{Introduction}
\vspace{-0.5mm}
\label{introduction}


Current neural architecture search (NAS) methods mainly include reinforcement learning-based NAS \cite{baker2016,zoph}, evolution-based NAS \cite{real2017large,liu2017hierarchical}, Bayesian optimization-based NAS \cite{kandasamy2018neural,zhou19e}, and gradient-based NAS \cite{luo2018neural,liu2018darts}, some of which have successfully been applied to related tasks for better architectures, such as semantic segmentation \cite{chen2018searching,liu2019auto} and object detection \cite{peng2019efficient,chen2019detnas,ghiasi2019fpn,tan2019efficientdet}. 

Among the NAS methods, gradient-based algorithms gain much attention because of the simplicity. Liu \emph{et al.} first propose the differentiable search framework, DARTS \cite{liu2018darts}, based on continuous relaxation and weight sharing \cite{pham2018efficient}, and inspire the follow-up studies \cite{xie2018snas,cai2018proxylessnas,chang2019data,xu2019pc,chen2019progressive}. In DARTS, different architectures share their weights as sub-graphs of a super-net. The super-net is trained for search, after which a target-net is derived for evaluation by manually keeping the important paths according to their softmax activations. Despite the simplicity, the architecture for evaluation only covers a small subset of the one for search, which causes a significant gap between super-net and target-net. We point out that the gap causes the following problems:

\begin{figure*}
	\centering
	\includegraphics[width=0.95\linewidth]{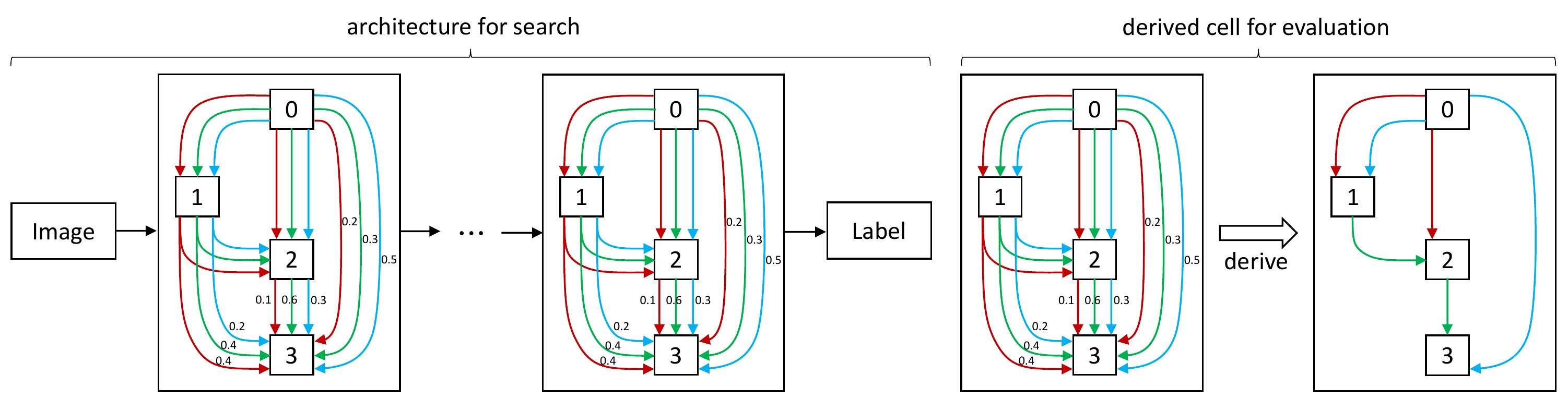}
	\caption{A diagram of DARTS. The target-net is derived by keeping the top-2 strongest connections of each node and has a significant gap with the architecture in search. The connections in different color represent candidate operations, with exemplar weights beside them.}
	\label{darts}
	\vspace{-3mm}
\end{figure*}

\begin{itemize}[leftmargin=20pt]
	\vspace{-1.5mm}
	\item \emph{inconsistent}: The super-net trained in the search phase is a summation among all candidate connections with a trainable distribution induced by softmax. It essentially optimizes a feature combination, instead of feature selection, which is the real goal of architecture search. As noted by \cite{chang2019data,yao2020efficient}, operations may be highly correlated. Even if the weight of some connection is small, the corresponding path may be indispensable for the performance. So the target-net derived from a high-performance super-net is not ensured to be a good one \cite{sciuto2019evaluating,yang2020ista}. The search process is inconsistent. 
	\vspace{-1.5mm}
	\item \emph{inefficient}: Because the super-net is a combination among all candidate connections, the whole graph needs to be stored in both forward and backward stages, which requires much memory and computational consumption. As a result, the search can be performed only on a very limited number of candidate operations, and the super-net is inefficient to train.
	\vspace{-1.5mm}
	\item \emph{inflexible}: The gap between the architectures in search and evaluation does not allow the search for topology in a differentiable way. In current methods \cite{liu2018darts,xie2018snas,cai2018proxylessnas,xu2019pc,chen2019progressive}, the target-net is derived based on a hand-crafted rule where each intermediate node keeps the top-2 strongest connections to prior nodes. However, there is no theoretical or experimental evidence showing that this rule is optimal. It limits the diversity of derived architectures in the topological sense \cite{huang2020explicitly}. Therefore, the search result is not flexible as we have no access to other kinds of topologies. 
\end{itemize}

Some studies adopt the Gumbel Softmax strategy \cite{jang2016categorical,maddison2016concrete} to sample a target-net that approaches to the one in search so that the gap can be reduced \cite{xie2018snas,wu2019fbnet,chang2019data,dong2019searching}. But still, the demand for computation and memory of the whole graph is not relieved. Chen \emph{et al.} \cite{chen2019progressive} propose a progressive shrinking method to bridge the depth gap between the super-net and target-net. NASP \cite{yao2020efficient} and ProxylessNAS \cite{cai2018proxylessnas} only propagate the proximal or sampled paths in search, which effectively reduces the computational cost. A recent study \cite{yang2020ista} relies on sparse coding to improve consistency and efficiency. However, all these methods do not support the search for flexible topologies in a differentiable way. DenseNAS \cite{fang2019densely} and PC-DARTS \cite{xu2019pc} introduce another set of trainable parameters to model path probabilities, but the target-net is still derived based on a hand-crafted rule.


In this paper, we aim to close the gap between the architectures in search and evaluation, and solve the problems mentioned above. Inspired by the observation that only one cell armed with learnable architecture parameters suffices to enable differentiable search, we introduce EnTranNAS composed of \underline{\textbf{En}}gine-cells and \underline{\textbf{Tran}}sit-cells. The Engine-cell is differentiable for architecture search as an \emph{engine}, while the Transit-cell only \emph{transits} the derived architecture. So the network in search is close to that in evaluation. We adopt a feature sharing strategy for more balanced parameter training of Transit-cell. It also reduces the computation and memory cost in search. Given that Engine-cell still has a gap with the derived architecture, we further develop an architecture derivation method that enables differentiable sparsification. The connections with non-zero weights are active for evaluation, which keeps the derived architecture equivalent to the one in search, and meanwhile supports the differentiable search for flexible topologies. 

We list the contributions of this study as follows:
\begin{itemize}[leftmargin=20pt]
	\vspace{-1mm}
	\item We propose a new NAS method, named EnTranNAS, which effectively reduces the gap between the architectures in search and evaluation. A feature sharing strategy is adopted for more balanced and efficient training of the super-net in search. 
	\vspace{-1mm}
	\item We develop a new architecture derivation method to replace the hand-crafted rule widely adopted in studies. The derived target-net has an equivalent architecture to the one in search, which closes the architecture gap between search and evaluation. It also makes topology learnable to explore more flexible search results. 
	\vspace{-1mm}
	\item Extensive experiments verify the validity of our proposed methods. We achieve an error rate of 2.22\% on CIFAR-10 with 0.07 GPU-day. Our method is able to efficiently search for flexible architectures of different scales directly on ImageNet and achieve a state-of-the-art top-1 error rate of 23.8\% in 2.1 GPU-day. 
\end{itemize}

\begin{figure*}
	\centering
	\includegraphics[width=0.95\linewidth]{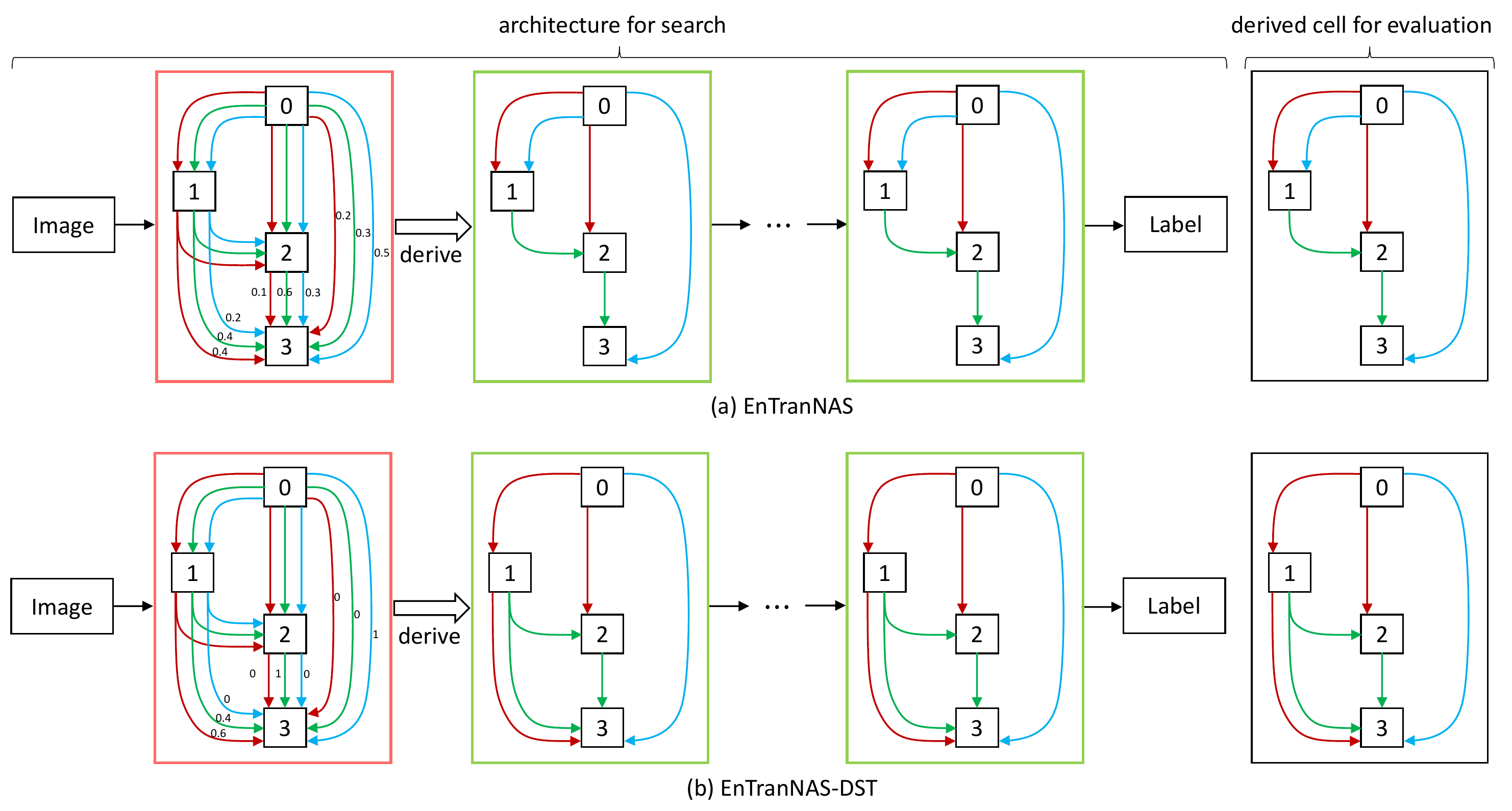}
	\caption{A diagram of our (a) EnTranNAS and (b) EnTranNAS-DST. Engine-cell and Transit-cell are in red and green boxes, respectively. EnTranNAS reduces the gap between the super-net and target-net. EnTranNAS-DST derive the architecture by keeping the connections with non-zero weights, so the valid computation graph in search is equivalent to the one of derived architecture in evaluation, and is not subject to any hand-crafted topology. The consistency is further improved and a flexible topology is supported. Zoom in to view better.}
	\label{f2}
	\vspace{-2mm}
\end{figure*}

\section{Methods}
In this section, we first briefly review the gradient-based search method widely adopted in current studies, and then develop our proposed methods, EnTranNAS and EnTranNAS-DST, respectively, showing that how they work to improve the consistency, efficiency, and flexibility of differentiable neural architecture search.


\subsection{Preliminaries}
In \cite{liu2018darts,xie2018snas,cai2018proxylessnas,chen2019progressive,xu2019pc,chang2019data}, the cell-based search space is represented by a directed acyclic graph (DAG) composed of $n$ nodes $\{x_1,x_2,\cdots,x_n\}$ and a set of edges $E=\{e^{(i,j)}|1\le i<j\le n\}$. For each edge $e^{(i,j)}$, there are $K$ connections in accordance with the candidate operations $\mathcal{O}=\{o_1,\cdots,o_K\}$. The forward propagation of the super-net for search is formulated as:
\begin{equation}
x_j=\sum_{i<j}\sum^K_{k=1}p^{(i,j)}_ko_k(w^{(i,j)}_k,x_i),
\end{equation}
where $p^{(i,j)}_k\in\{0,1\}$ is a binary variable that indicates whether the connection is active, $o_k$ denotes the $k$-th operation, and $w^{(i,j)}_k$ is its corresponding weight on this connection and becomes none for non-parametric operations, such as max pooling and identity. Since binary variables are not easy to optimize in a differentiable way, continuous relaxation is adopted and $p^{(i,j)}_k$ is relaxed as: 
\begin{equation}
p^{(i,j)}_k=\frac{\exp(\alpha^{(i,j)}_k)}{\sum_k\exp(\alpha^{(i,j)}_k)},
\end{equation}
where $\alpha^{(i,j)}_k$ is the introduced architecture parameter jointly optimized with the super-net weights. After search, as shown in Figure~\ref{darts}, a target-net is derived according to a hand-crafted rule based on $p^{(i,j)}_k$ as the importance of connections. We let $\mathbf{P}\in R^{|E|\times K}$ denote the matrix formed by $p^{(i,j)}_k$, and the forward propagation of the target-net for evaluation is formulated as:
\begin{equation}
x_j=\sum_{(i,k)\in S_j}o_k(w^{(i,j)}_k,x_i),
\label{eval}
\end{equation}
\begin{equation}
S_j=\{(i,k)|A^{(i,j)}_k=1,\forall i<j, 1\le k\le K\},
\label{handcrafted}
\end{equation}
where $A^{(i,j)}_k$ is the element of $\mathbf{A}\in \{0,1\}^{|E|\times K}$ and we have $\mathbf{A}=\mathtt{Proj}_{\Omega}(\mathbf{P})$, where $\Omega$ denotes the hand-crafted rule by which only the top-2 strongest elements of each node $j$ in $\mathbf{P}$ are projected onto $1$ and others are $0$. 

It is shown that there is a gap between the super-net and target-net in DARTS. As mentioned in Section \ref{introduction}, the gap may cause inconsistency with target-net, and the super-net is inefficient to train. Besides, the hand-crafted rule restricts the derived architecture to a fixed topology.

\subsection{Engine-cell and Transit-cell}


Given that only one cell armed with learnable parameters suffices to enable differentiable search, we aim to re-design the DARTS framework. First, at the super-net level, we introduce EnTranNAS composed of \underline{\textbf{En}}gine-cells and \underline{\textbf{Tran}}sit-cells. As shown in Figure~\ref{f2} (a), the architecture derivation is not a post-processing step as in DARTS, but is performed at each iteration of search. Engine-cell has the same role as the cell in DARTS and stores the whole DAG. It performs architecture search as an \emph{engine} by optimizing architecture parameters $\alpha^{(i,j)}_k$. As a comparison, Transit-cell only \emph{transits} the currently derived architecture as a sub-graph into later cells. By doing so, EnTranNAS keeps the differentiability for architecture search by Engine-cell, and effectively reduces the gap between the super-net and target-net using Transit-cells. At the final layer of super-net, representation is output from a Transit-cell, which has the same architecture as the target-net. Thus, with more confidence, a higher super-net performance indicates a better target-net architecture. Besides, a huge amount of computation and memory overhead in Transit-cells is saved. We can accordingly use a larger batchsize to speed up the search process, or adopt a larger search space with more candidate operations due to the memory relief. 

By introducing a temperature parameter \cite{xie2018snas,wu2019fbnet}, we calculate $p^{(i,j)}_k$ in Engine-cell as:
\begin{equation}
p^{(i,j)}_k=\frac{\exp(\alpha^{(i,j)}_k/\tau)}{\sum_k\exp(\alpha^{(i,j)}_k/\tau)},
\label{pk}
\end{equation}
where $\tau$ is a temperature parameter. As $\tau\rightarrow0$, $p^{(i,j)}_k$ approaches to a one-hot vector. We do not introduce the Gumble random variables as adopted in \cite{xie2018snas,wu2019fbnet} because our architecture is not derived by sampling. We anneal $\tau$ with epoch so that Engine-cell approximately performs feature selection after convergence and can be close to the derived architecture in Transit-cell. 

\subsection{Feature Sharing Strategy}

Since Transit-cell only conducts the derived sub-graph, only a small portion of super-net weights $w^{(i,j)}_k$ is optimized in Transit-cell at each update. It impedes the training efficiency of super-net and may affect the search result due to the uneven optimization on candidate operations. In order to circumvent this issue and have a balanced parameter training for Transit-cells, we introduce a feature sharing strategy within the cell level. 

We notice that the non-parametric operation from a node to different nodes always produces the same features, which can be stored and computed only once. We extend it to parameterized operations, by making the simplification that the same operation from node $i$ to other nodes $j>i$ always shares the same feature in one cell. The output of node $x_j$ in our EnTranNAS is thus formulated as:
\begin{eqnarray}\label{feature_sharing}
x_j=\left\{
\begin{array}{ll}
\sum_{i<j}\sum^K_{k=1}p^{(i,j)}_ko_k(w^{(i)}_k,x_i), & \text{in Engine-cell},\\
\sum_{(i,k)\in S_j}o_k(w^{(i)}_k,x_i), & \text{in Transit-cell},
\end{array}
\right.
\end{eqnarray}
where $w^{(i)}_k$ is the parameter of operation $k$ for node $i$, and becomes none for non-parametric operations. In this way, the number of trainable connections in one cell is reduced from $|E|\times \bar{K}$ to $(n-1)\times \bar{K}$, where $\bar{K}$ denotes the number of parametrized operations and $|E|=C^2_n$. Consequently, the less learnable parameters have a more balanced opportunity to be optimized. In addition, the feature of one operation from the node $i$ is calculated only once and is re-used for later nodes $j>i$ in the same cell, which saves much computation and memory overhead and accelerates the search. Note that the feature sharing strategy harms the representation power of super-net. However, it does not affect the search validity as the features for selection are still produced by the same operations on the same nodes. What we search for is which operation performed on which node, instead of how their parameters are optimized. 




\subsection{Differentiable Search for Topology}

\begin{figure*}
	\centering
	\includegraphics[width=0.98\linewidth]{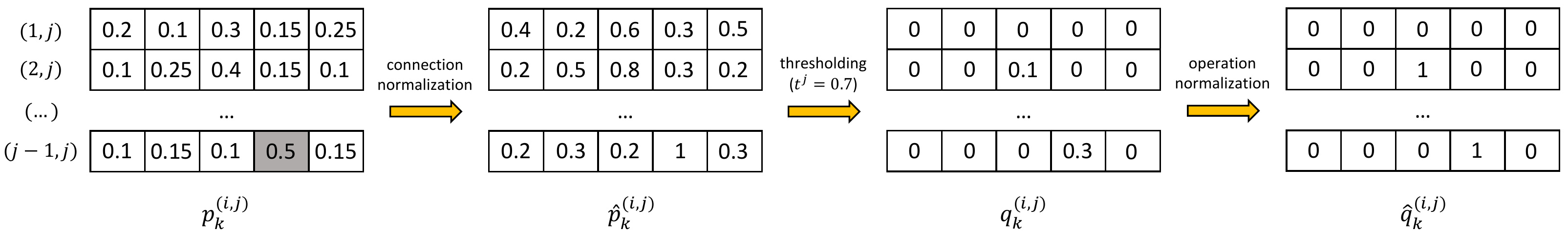}
	\caption{An illustration of the computation procedures of $\hat{q}^{(i,j)}_k$ as an example. The gray bin denotes the maximal element of $p^{(i,j)}_k$ for all $1\le k\le K$ and $i<j$. There is at least one connection left for each intermediate node $j$ since $\max_{i<j,1\le k\le K}\{\hat{p}^{(i,j)}_k\}=1$ and $t^{(j)}<1$.}
	\label{f3}
	\vspace{-1mm}
\end{figure*}

\begin{figure*}[!t]
	\centering
	\includegraphics[width=0.95\linewidth]{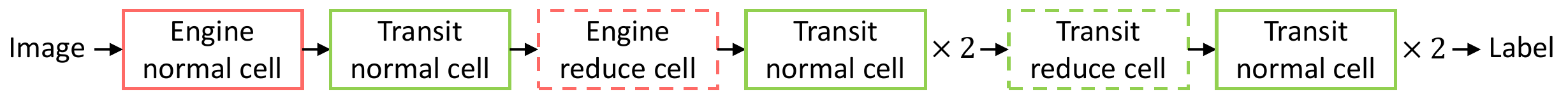}
	\caption{Our architecture for search. Engine-cell and Transit-cell are shown in red and green boxes, respectively. Normal and reduction cells are shown in solid and dotted boxes, respectively.}
	\label{cell_dist}
	\vspace{-2mm}
\end{figure*}

Albeit EnTranNAS reduces the gap between super-net and target-net, the Engine-cell computes the whole graph and is still different from the derived cell for evaluation. To this end, we further reduce the gap by proposing a new architecture derivation method that supports differentiable sparsification and enables the search for topology, named EnTranNAS-DST. As shown in Figure~\ref{f2} (b), in Engine-cell, the non-derived connections always have zero weights, such that the valid propagation of Engine-cell is equivalent to that of the derived cell, which eliminates the gap between the architectures in search and evaluation.

In prior studies \cite{liu2018darts,chen2019progressive,xu2019pc}, connection coefficients are induced as softmax activations and thus do not support zero values. A differentiable sparsification method is proposed in \cite{lee2019differentiable} for network pruning. We combine both advantages to keep the softmax activations and also enable the differentiability for zero weights. Concretely, since we need to cut out connections for each intermediate node instead of edge, we compute $p^{(i,j)}_k$ by Eq.~(\ref{pk}), and then perform a connection normalization for each intermediate node $j>1$ as:
\begin{equation}
\hat{p}^{(i,j)}_k=\frac{p^{(i,j)}_k}{\max\limits_{i<j,1\le k\le K}\{p^{(i,j)}_k\}},
\end{equation}
where $\hat{p}^{(i,j)}_k$ is the activation after connection normalization. We introduce another set of trainable parameters $\{\beta^{(j)}\}^{n}_{j=2}$ and have the threshold of each intermediate node by $t^{(j)}=\text{sigmoid}(\beta^{(j)})$. With the thresholds, we can prune these connections as: 
\begin{equation}
q^{(i,j)}_k=\sigma(\hat{p}^{(i,j)}_k-t^{(j)}),
\label{relu}
\end{equation}
where $\sigma$ denotes the ReLU function. Finally, if there exists a $k$ such that $q^{(i,j)}_k\ne0$ for edge $(i,j)$, we perform an operation normalization by:
\begin{equation}
\hat{q}^{(i,j)}_k=\frac{q^{(i,j)}_k}{\sum_k{q^{(i,j)}_k}},
\end{equation}
where $\hat{q}^{(i,j)}_k$ is used as the coefficients of connections. It enables sparsification in a differentiable way. Given that $\max_{i<j,1\le k\le K}\{\hat{p}^{(i,j)}_k\}=1$ and $t^{(j)}<1$, there is at least one connection left for each intermediate node $j$ by Eq.~(\ref{relu}), so the cell structure will not be broken, and will keep valid along the training. An illustration of how do we compute $\hat{q}^{(i,j)}_k$ is shown in Figure~\ref{f3}.

In Engine-cell, we replace the $p^{(i,j)}_k$ in Eq.~(\ref{feature_sharing}) with $\hat{q}^{(i,j)}_k$ for search. To derive the architecture in Transit-cell or for evaluation, the $S_j$ in Eq.~(\ref{feature_sharing}) is changed from Eq.~(\ref{handcrafted}) to the following form:
\begin{equation}
S_j=\{(i,k)|\hat{q}^{(i,j)}_k>0,\forall i<j, 1\le k\le K\},
\end{equation}
by which we only keep the connections with non-zero coefficients as the derived architecture, which eliminates its gap with the super-net architecture, and meanwhile does not restrict the architecture to any fixed topology.

In implementation, we enforce sparsification by adding a regularization. Our optimization objective is in accordance with the bi-level manner introduced in \cite{liu2018darts}. The upper-level loss function of our super-net when optimizing the architecture parameters $\{\alpha^{(i,j)}_k\}$ and $\{\beta^{(j)}\}$ is formulated as: 
\begin{equation}
\min_{\alpha,\beta}\quad \mathcal{L}_{val}\left(\alpha,w^{*}\right)+\lambda\frac{1}{n-1}\sum\nolimits^n_{j=2}-\log(t^{(j)}),
\label{eq11}
\end{equation}
where $\mathcal{L}_{val}\left(\alpha,w^{*}\right)$ is the validation loss with the current network parameters $w^{*}$, and $\lambda$ is a hyper-parameter by which we can control the degree of sparsification to obtain more flexible topologies. We visualize our search process of EnTranNAS-DST ($\lambda=0.1$) in the video attached in the supplementary material. Its corresponding description is shown in the Appendix file.

\subsection{Implementations}

For both EnTranNAS and EnTranNAS-DST, we set the first normal and reduction cells as Engine-cells, and the other cells as Transit-cells. The super-net with 8 cells for search on CIFAR-10 is shown in Figure~\ref{cell_dist}. The first cells of normal and reduction cells are set as Engine-cells, while the others are Transit-cells. In experiments, we compare it with other configurations in Table~\ref{acc} to ablate our design choice. 

Similar to the partial channel connection strategy in \cite{xu2019pc}, we also try to reduce the number of channels to further save memory cost and reduce search time. Different from their method, we adopt the bottleneck technique that is popular in manually designed networks \cite{he2015resbet,huang2016densely}. Concretely, we perform a $1\times1$ convolution to reduce the number of channels by a ratio before feeding a node into all candidate operations. Another $1\times1$ convolution is appended to recover the number of channels to form each intermediate node. The reduction ratio is set as 4 in our experiments. 

\section{Related Work}
\vspace{-0.5mm}

Reinforcement learning is first adopted to assign the better architecture with a higher reward in \cite{baker2016,zoph}. Follow-up studies focus on reducing the computational cost \cite{Zoph_2018_CVPR,Zhong_2018_CVPR,liu2018progressive,cai2018efficient,pmlr-v80-cai18a,pham2018efficient}. As another line of NAS methods, evolution-based algorithms search for architectures as an evolving process towards better performance \cite{xie2017genetic,real2017large,liu2017hierarchical,real2019regularized,elsken2018efficient,miikkulainen2019evolving}. However, the search cost of both reinforcement leaning- and evolution-based methods is still demanding for practical applications. A good solution to this problem is one-shot methods that constructs a super-net covering all candidate architectures \cite{bender2018understanding,brock2017smash}. The super-net is trained only once in search and is then deemed as a performance estimator. Some studies train the super-net by sampling a single path \cite{guo2019single,li2019random,you2020greedynas} in a chain-based search space \cite{hu2020dsnas,cai2019once,mei2019atomnas,yu2020bignas}. As a comparison, DARTS-based methods \cite{liu2018darts,xie2018snas} introduce architecture parameters jointly optimized with the super-net weights and performs the differentiable search in a cell-based space. Our study belongs to this category because it enables to discover more complex connecting patterns. 

Despite the simplicity of DARTS, the architecture gap between search and evaluation impedes its validity. Follow-up studies aim to reduce the gap \cite{xie2018snas,chen2019progressive,chang2019data,yang2020ista}, improve the search efficiency \cite{yao2020efficient,yang2020ista}, and model path probabilities \cite{xu2019pc}. However, all these methods derive the final architecture based on a hand-crafted rule, which inevitably limits the topology. Our method differs from these studies in that the super-net of EnTranNAS-DST dynamically changes in the search phase in a differentiable way, and then derives a target-net that has the same architecture as the one in search, and is not subject to any specific topology.

\begin{table*}
	\begin{minipage}[th!]{\textwidth}
		\renewcommand\arraystretch{1.15}
		\setlength\tabcolsep{1.3mm}
		\begin{minipage}[t]{0.25\textwidth}
			\begin{center}
				\small
				\begin{tabular}{c|c|c}
					\hline
					\multirow{2}*{Engine-cell} & Super-net & Child-net \\
					~ & Acc. (\%) & Acc. (\%)\\
					\hline
					all (DARTS) & 88.29 & 63.97 \\
					\hline
					one half & 87.45 & 65.51 \\
					\hline
					last & 84.02 & 83.35 \\
					\hline
					first & 86.68 & 86.24 \\
					\hline
				\end{tabular}
			\end{center}
			\vspace{-1mm}
			\caption{Super-net accuracy drop in different settings of Engine-cell.}
			\label{acc}
		\end{minipage}
		\quad\quad\ 
		\begin{minipage}[t]{0.21\textwidth}
			\begin{center}
				\small
				\begin{tabular}{l|c}
					\hline
					Methods & Kendall $\tau$ \\
					\hline
					DARTS & -0.47 \\
					P-DARTS & 0.20 \\
					PC-DARTS & -0.07 \\
					\hline
					EnTranNAS & 0.33 \\
					EnTranNAS-DST & 0.60 \\
					\hline
				\end{tabular}
			\end{center}
			\vspace{-1mm}
			\caption{Kendall scores of our and existed methods.}
			\label{kendall}
		\end{minipage}	
		\quad\quad\ 
		\begin{minipage}[t]{0.4\textwidth}
			\begin{center}
				\small
				\begin{tabular}{l|c|c|c}
					\hline
					~ & Memory & Batchsize & Cost\\
					~ & (G) & (64) & (GPU-day)\\
					\hline
					DARTS (1st order) & 9.0 & $\times$1 & 0.73\\
					\hline
					+Engine\&Transit-cell & 4.5 & $\times$2 & 0.22\\
					\hline
					+feature sharing & 2.6 & $\times$4 & 0.09\\
					\hline
					+bottleneck & 1.5 & $\times$8 & 0.06\\
					\hline
				\end{tabular}
			\end{center}
			\vspace{-1mm}
			\caption{Efficiency improved by each component. The three components are accumulated from top to bottom.}
			\label{eff}
		\end{minipage}
	\end{minipage}
\end{table*}




\begin{figtab*}[t]
	\begin{minipage}[t!]{0.4\linewidth}
		\renewcommand\arraystretch{1.1}
		\setlength\tabcolsep{1.8mm}
		\begin{center}
			\begin{tabular}{l|c|c|c}
				\hline
				\multirow{2}*{$\lambda$} & Edges & Params & Flops\\
				~ & (N / R) & (M) & (M)\\
				\hline
				0.2 & 9 / 8 & 5.07 & 580\\
				\hline
				0.1 & 11 / 6 & 5.88 & 673\\
				\hline
				0.05 & 13 / 14 & 6.99 & 779\\
				\hline
			\end{tabular}
		\end{center}
		\vspace{-1mm}
		\tabcaption{EnTranNAS-DST with different $\lambda$. ``N'' and ``R'' denote normal and reduction cell, respectively. It is shown that the number of edges is not fixed to access flexible topologies with variant capacities.}
		\label{flex}
	\end{minipage}\quad\ \
	\begin{minipage}[t!]{0.57\linewidth}
		\centering
		\vspace{-1mm}
		\includegraphics[width=1.\linewidth]{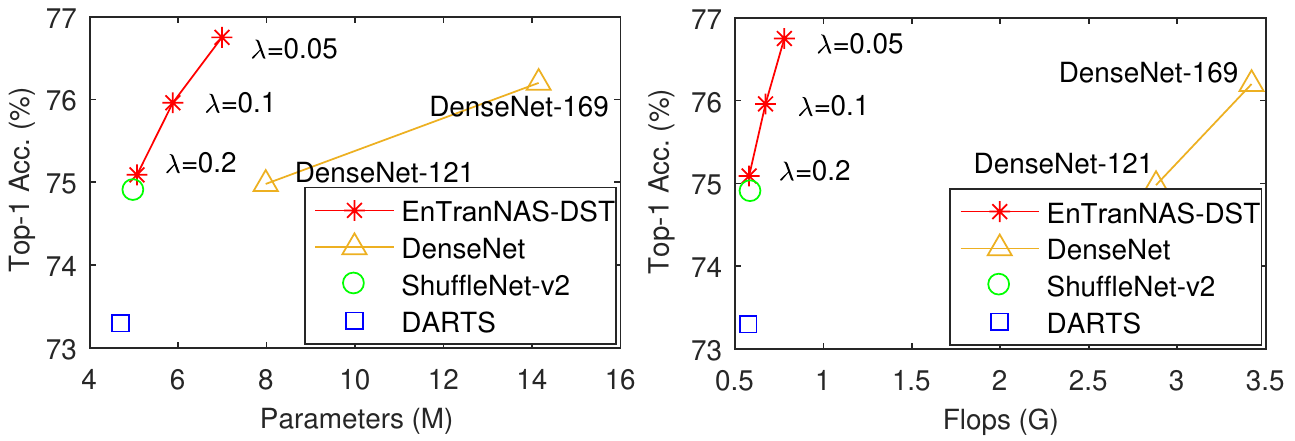}
		\figcaption{Comparison of top-1 accuracies on ImageNet with parameters (left) and Flops (right). Zoom in to view better.}
		\label{curve}
	\end{minipage}
\end{figtab*}

\begin{table*}[!t]
	\footnotesize
	\begin{center}
		\begin{tabular}{l|c|c|c|c|c|c}
			\hline
			\multirow{2}*{Backbone} & \multirow{2}*{ResNet-50 \cite{he2015resbet}} & \multirow{2}*{NASNet-A \cite{Zoph_2018_CVPR}} & \multirow{2}*{DARTS \cite{liu2018darts}} & EnTranNAS-DST & EnTranNAS-DST & EnTranNAS-DST \\
			~  & ~ & ~ & ~ & ($\lambda=0.2$) & ($\lambda=0.1$) & ($\lambda=0.05$)\\
			\hline
			mIoU (\%) & 76.5 & 75.4 & 75.1 & 76.3 & 76.8 & 77.1\\
			\hline
		\end{tabular}
	\end{center}
	\vspace{-2mm}
	\caption{Results of semantic segmentation on Pascal VOC 2012 using different architectures as the backbone with the same DeepLabV3 head \cite{chen2017rethinking}, input size of $513\times513$, and output stride of 16 in the single scale inference setting.}
	\vspace{-2mm}
	\label{semantic}
\end{table*}

\section{Experiments}

We first analyze how each of our designs improves the consistency, efficiency and flexibility by ablation studies, and then compare our results on CIFAR-10 and ImageNet with state-of-the-art methods. \textbf{All our searched architectures are visualized in the Appendix file.}

\subsection{Ablation Studies}

\textbf{Consistency.} EnTranNAS reduces the gap between the super-net and target-net. We test the effects of our design with different settings. After search on CIFAR-10, we perform inference only through the paths in the derived architecture as a child-net and compare their validation accuracy changes. As shown in Table~\ref{acc}, when all cells are set as Engine-cell, the super-net is equivalent to DARTS and has the largest accuracy drop. Making one half of cells as Engine-cell also causes a large accuracy drop. As a comparison, when one Engine-cell is used, we have a small accuracy drop, which demonstrates the validity of our method to reduce the gap. We set the first cell as Engine-cell because it relatively has a better super-net accuracy and a smaller accuracy drop than the last cell setting. 

We also adopt the Kendall metric \cite{kendall1938new} that evaluates the rank correlation of data pairs. It ranges from -1 to 1 as the ranking order changes from being reversed to identical. We run DARTS, P-DARTS, PC-DARTS, EnTranNAS and EnTranNAS-DST on CIFAR-10 for six times with different seeds, and retrain these searched architectures. We calculate the Kendall metric for each method using the six retrained and super-net accuracies in Table~\ref{kendall}. It is shown that our methods help to improve the consistency.

\textbf{Efficiency.}  The improved efficiency of our search on CIFAR-10 by each component is shown in Table~\ref{eff}. ``Memory'' shows the memory consumption with a batchsize of 64. ``Batchsize'' is the largest batchsize that can be used on a single GTX 1080 Ti GPU. ``Cost'' denotes the corresponding search time using the enlarged batchsize. Both of our Engine\&Transit-cell design and feature sharing strategy significantly improve the search efficiency. Similar to \cite{xu2019pc} that reduces the number of channels when performing all operations, we adopt a bottleneck before operations. When ``bottleneck'' is added, we can use a batchsize of 512 and reduce the search time to 0.06 GPU-day, which is about ten times as fast as our re-implementation of DARTS. 

\begin{table*}[!t]
	\renewcommand\arraystretch{1.12}
	\setlength\tabcolsep{3.5mm}
	\begin{center}
		\begin{tabular}{lcccc}
			\hline
			\multirow{2}*{Methods} & Test Error  & Params & Search Cost  & \multirow{2}*{Search Method}\\
			~  & (\%) &  (M) & (GPU-day) & ~\\
			\hline
			\hline
			DenseNet-BC \cite{huang2016densely} & 3.46 & 25.6 & - & manual\\
			\hline
			NASNet-A + cutout \cite{Zoph_2018_CVPR} & 2.65 & 3.3  & 1800 & RL\\
			ENAS + cutout \cite{pham2018efficient} & 2.89 & 4.6 & 0.5  & RL\\
			AmoebaNet-B +cutout \cite{real2019regularized} & 2.55$\pm$0.05 & 2.8 & 3150 & evolution \\
			Hierarchical Evolution \cite{liu2017hierarchical} & 3.75$\pm$0.12 & 15.7 & 300 & evolution\\
			\hline
			DARTS (2nd order) + cutout \cite{liu2018darts} & 2.76$\pm$0.09 & 3.3 & 4.0 & gradient \\
			SNAS (moderate) + cutout \cite{xie2018snas} & 2.85$\pm$0.02 & 2.8 & 1.5 & gradient \\
			ProxylessNAS+cutout \cite{cai2018proxylessnas} & 2.08$^{\dag}$ & 5.7 & 4.0 & gradient\\ 
			PC-DARTS + cutout \cite{xu2019pc} & 2.57$\pm$0.07 & 3.6 & 0.1 & gradient\\ 			
			NASP + cutout \cite{yao2020efficient} & 2.83$\pm$0.09 & 3.3 & 0.1 & gradient\\
			MiLeNAS + cutout \cite{he2020milenas} & \textbf{2.51$\pm$0.11} & 3.87 & 0.3 & gradient\\
			\hline
			EnTranNAS + cutout & {2.53$\pm$0.06} & 3.45 & \color{blue}{\textbf{0.06}} & gradient\\
			EnTranNAS-DST + cutout & \color{blue}{\textbf{2.48$\pm$0.08}} & 3.20 & \textbf{0.10} & gradient\\
			\hline
			\hline
			NASP (12 operations) + cutout \cite{yao2020efficient} & 2.44$\pm$0.04 & 7.4 & 0.2 & gradient\\ 				
			EnTranNAS (12 operations) + cutout & \textbf{\color{blue}{2.22$\pm$0.05}} & 7.68 & \textbf{0.07} & gradient\\
			\hline
			
		\end{tabular}
	\end{center}
	\caption{Search results on CIFAR-10 and comparison with state-of-the-art methods. Search cost is tested on a single NVIDIA GTX 1080 Ti GPU. The best and second best results are shown in blue and black bold. Methods with the notation ``(12 operations)'' search on an extended search space with 12 operations. \dag: ProxylessNAS uses a different macro-architecture from the other methods.}
	\label{tab1}
	\vspace{-1mm}
\end{table*}


\textbf{Flexibility.} EnTranNAS-DST enables the differentiable search for topology and does not limit the number of edges in normal or reduction cells. We can obtain architectures with diverse capacities. A larger $\lambda$ makes $t^{(j)}$ closer to 1, which cuts out more connections by Eq.~(\ref{relu}) and leads to a more sparse architecture. Our search results on ImageNet with different $\lambda$ are shown in Table~\ref{flex}. Their accuracies on ImageNet validation are depicted as a function of parameters and FLOPs in Figure~\ref{curve}. It is shown that we have a better trade-off than the strong baseline of manually designed architecture, DenseNet. Our EnTranNAS-DST ($\lambda$=0.05) surpasses DenseNet-169 with about one half of parameters and less than one fourth of FLOPs. We also transfer these searched architectures to semantic segmentation in Table~\ref{semantic}, which shows that our architectures with diverse capacities are also applicable to other tasks. Our method breaks the topology constraint and enables to search for flexible results even outside the mobile setting limitation, which is beyond the ability of most existed NAS methods and extends the potential applications of searched architectures. 





\subsection{Results on CIFAR-10}
We describe the CIFAR-10 dataset in the Appendix file. The super-net for search on CIFAR-10 is composed of 8 cells (6 normal cells and 2 reduction cells) with the initial number of channels as 16. There are 6 nodes in each cell. The first 2 nodes in cell $k$ are input nodes, which are the outputs of cell $k-2$ and $k-1$, respectively. The output of each cell is the concatenation of all intermediate nodes. We train the super-net for 50 epochs with a batchsize of 512. SGD is used to optimize the super-net weights with a momentum of 0.9 and a weight decay of 3e-4. Its learning rate is set as 0.2 and is annealed down to zero with a cosine scheduler. We use the Adam optimizer for the architecture parameters $\{\alpha^{(i,j)}_k\}$ (and $\{\beta^{(j)}\}$ for EnTranNAS-DST) with a learning rate of 6e-4, a momentum of (0.5, 0.999) and a weight decay of 1e-3. The initial temperature in Eq.~(\ref{pk}) is set as 5.0 and is annealed by 0.923 every epoch. We run our search for 5 times and choose the architecture with the best validation accuracy as the searched one. In evaluation, the target-net has 20 cells (18 normal cells and 2 reduction cells) with the initial number of channel as 36. We train for 600 epochs with a batchsize of 96, and report the mean error rate with the standard deviation of 5 independent runs. SGD optimizer is used with a momentum of 0.9, a weight decay of 3e-4, and a gradient clipping of 5. The initial learning rate is set as 0.025 and is annealed down to zero following a cosine scheduler. As convention, a cutout length of 16, a drop out rate of 0.2, and an auxiliary head are adopted. 

We search on CIFAR-10 from the standard and extended version of candidate operation space. The standard space has 7 operations and is consistent with current studies \cite{liu2018darts,xie2018snas,chen2019progressive,xu2019pc}. The extended version additionally has 5 more operations, which are $1\times1$ convolution, $3\times3$ convolution, $1\times3$ then $3\times1$ convolution, $1\times5$ then $5\times1$ convolution, and $1\times7$ then $7\times1$ convolution. The two versions are listed in the Appendix file. As shown in Table~\ref{tab1}, for the standard search space, EnTranNAS achieves a state-of-the-art performance of 2.53\% error rate with only 0.06 GPU-day. The accuracy is on par with MiLeNAS \cite{he2020milenas}, whose search cost is 5 times as much as ours. To our best knowledge, 0.06 GPU-day is the top speed on DARTS-based search space. EnTranNAS-DST achieves a better performance with less parameters than EnTranNAS due to its superiority in learnable topology. When we search on the extended search space, a higher-performance architecture is searched with an error rate of 2.22\%, which is better than NASP \cite{yao2020efficient} that also searches on 12 operations. The search cost still has superiority and is increased by only 0.01 GPU-day than that on the standard version. That is because the extra operations only add the computational cost on Engine-cells, which account for a small portion of the super-net in search. Therefore, the search cost of EnTranNAS increases sub-linearly as the search space is enlarged.


\begin{table*}
	\renewcommand\arraystretch{1.12}
	\setlength\tabcolsep{3.mm}
	\begin{center}
	\begin{tabular}{lcccccc}
		\hline
		\multirow{2}*{Methods} & \multicolumn{2}{c}{Test Err. (\%)} & Params & Flops & Search Cost & \multirow{2}*{Search Method}\\
		\cline{2-3}
		~ & top-1 & top-5 &  (M) & (M) & (GPU days) & ~\\
		\hline
		\hline
		Inception-v1 \cite{Szegedy2015} & 30.2 & 10.1 & 6.6 & 1448 & - & manual \\
		MobileNet \cite{howard2017mobilenets} & 29.4 & 10.5 & 4.2 & 569 & - & manual\\
		ShuffleNet 2$\times$ (v2) \cite{ma2018shufflenet} & 25.1 & - & $\sim$5 & 591 & - & manual\\
		\hline
		MnasNet-92 \cite{tan2019mnasnet} & 25.2 & 8.0 & 4.4 & 388 & - & RL\\
		AmoebaNet-C \cite{real2019regularized} & 24.3 & 7.6 & 6.4 & 570 & 3150 & evolution\\
		\hline
		DARTS (2nd order) \cite{liu2018darts} & 26.7 & 8.7 & 4.7 & 574 & 4.0 & gradient\\
		SNAS \cite{xie2018snas} & 27.3 & 9.2 & 4.3 & 522 & 1.5 & gradient\\
		P-DARTS \cite{chen2019progressive} & 24.4 & 7.4 & 4.9 & 557 & 0.3 & gradient\\
		ProxylessNAS (ImageNet) \cite{cai2018proxylessnas} & 24.9 & 7.5 & 7.1 & 465 & 8.3 & gradient\\
		PC-DARTS (ImageNet) \cite{xu2019pc} & \textbf{24.2} & 7.3 & 5.3 & 597 & 3.8 & gradient\\
		\hline
		EnTranNAS (CIFAR-10) & 24.8 & 7.6 & 4.9 & 562 & 0.06 & gradient\\
		EnTranNAS (ImageNet) & 24.3 & \textbf{7.2} & 5.5 & 637 & \color{blue}{\textbf{1.9}} & gradient\\
		\hline			
		EnTranNAS-DST (ImageNet) $^{\dag}$ & \color{blue}{\textbf{23.8}} & \color{blue}{\textbf{7.0}} & 5.2 & 594 & \textbf{2.1} & gradient\\
		\hline
	\end{tabular}
	\end{center}
	\caption{Search results on ImageNet and comparison with state-of-the-art methods. Search cost is tested on eight NVIDIA GTX 1080 Ti GPUs. ``(ImageNet)'' indicates the method is directly searched on ImageNet. Otherwise, it is searched on CIFAR-10, and then transfered to ImageNet. ${\dag}$: The result is searched with $\lambda$ as 0.2 under the mobile setting and selected out as the best from five implementations.}
	\label{tab2}
	\vspace{-1mm}
\end{table*}

\vspace{-1mm}
\subsection{Results on ImageNet}
\vspace{-0.5mm}
We describe the ImageNet dataset in the Appendix file. Following \cite{xu2019pc}, we perform three convolution layers of stride of 2 to reduce the resolution from the input size $224\times224$ to $28\times28$. The super-net for search has 8 cells with the initial number of channels as 16, while the target-net for evaluation has 14 cells and starts with 48 channels. We use a batchsize of 1,024 for both search and evaluation. In search, we train for 50 epochs with the same optimizers, momentum, and weight decay as that on CIFAR-10. The initial learning rate of network weights is 0.5 (annealed down to zero following a cosine scheduler). The learning rate of architecture parameters $\{\alpha^{(i,j)}_k\}$ (and $\{\beta^{(j)}\}$ for EnTranNAS-DST) is 6e-3. The initial temperature and its annealing ratio for EnTranNAS are the same as that on CIFAR-10. For EnTranNAS-DST, the initial temperature is set as 1 and is annealed by 0.9 every epoch. In evaluation, we train for 250 epochs from scratch using the SGD optimizer with a momentum of 0.9 and a weight decay of 3e-5. The initial learning rate is set as 0.5 and is annealed down to zero linearly. Following \cite{xu2019pc}, an auxiliary head and the label smoothing technique are also adopted. 

We use both EnTranNAS and EnTranNAS-DST for experiments on ImageNet with the standard search space. As shown in Table~\ref{tab2}, EnTranNAS searched on CIFAR-10 has a top-1 error rate of 24.8\%, which is competitive given that its search time is much more friendly than other methods. We also directly search on ImageNet. EnTranNAS achieves a top-1/5 error rates of 24.3\%/7.2\%, which is on par with PC-DARTS whose search cost is twice as much as ours. Different from other studies, EnTranNAS-DST is the only method that does not limit the topology of searched architecture. When $\lambda$ in Eq.~(\ref{eq11}) is 0.2, a model with less parameters and FLOPs is searched and has a top-1 error rate of 23.8\%, which surpasses EnTranNAS (ImageNet) by 0.5\% error rate due to its explicit learning of topology. The search cost is larger than EnTranNAS because at the beginning of search all connections to a node have non-zero weights and are kept active. As the search proceeds, EnTranNAS-DST adaptively drops connections. An illustration of how EnTranNAS-DST changes its derived architecture in search is shown in the supplementary video \footnote{\href{https://drive.google.com/file/d/1FFy7gv9uTkv2fEXVN1_vhzoijl-FWUDb/view?usp=sharing}{Google drive link}} and described in the Appendix file. We see its search is still faster than PC-DARTS but enjoys better performances and flexibilities. We show in our ablation studies that architectures with flexible topologies of diverse capacities can be searched by controlling the hyper-parameter $\lambda$.




\vspace{-1mm}
\section{Conclusion}
\vspace{-0.5mm}

In this paper, we introduce EnTranNAS that reduces the gap between the architectures in search and evaluation and saves much computational and memory cost. A feature sharing strategy is adopted for more efficient and balanced training of search. We further propose EnTranNAS-DST that closes the gap by a new architecture derivation method. It supports the search for flexible architectures without topology constraint. Experiments show that EnTranNAS improves the consistency and efficiency, and EnTranNAS-DST extends the flexibility of searched architectures. We produce state-of-the-art results on CIFAR-10 and directly on ImageNet with obvious superiority in search cost.  


\newpage

{\small
\bibliographystyle{ieee_fullname}
\bibliography{neural_arch}
}
\newpage

\appendix

\section*{Appendix A: Datasets and Search Spaces}
\label{dataset}
\subsection*{A.1 Datasets}
The CIFAR-10 dataset consists of 60,000 images of size $32\times32$ in 10 classes. There are 50,000 images for training and 10,000 images for testing. In search, we use a half of the training set to optimize network weights and the other half as the validation set to optimize architecture parameters. 

The ImageNet dataset contains 1.2 million training images, 50,000 validation images, and 100,000 test images. Following \cite{xu2019pc}, we directly perform the search on ImageNet by randomly sampling 10\% images of the training set for network weights, and another 10\% for architecture parameters.

The standard data augmentation methods are used for both CIFAR-10 and ImageNet.


\subsection*{A.2 Search Spaces}
\label{search_space}
Following \cite{liu2018darts,chen2019progressive,xu2019pc}, we search on the commonly used operation space that includes $3\times3$ and $5\times5$ separable convolution, $3\times3$ and $5\times5$ dilated separable convolution, $3\times3$ max pooling, $3\times3$ average pooling, identity, and zero. Because our method is friendly to memory consumption and thus can be performed on a larger search space, we adopt an extended version of operation space that has 5 extra operations: $1\times1$ convolution, $3\times3$ convolution, $1\times3$ then $3\times1$ convolution, $1\times5$ then $5\times1$ convolution, and $1\times7$ then $7\times1$ convolution. The two versions of operation space are listed in Table~\ref{operation}. The zero operation is used to indicate the lack of connection between two nodes \cite{liu2018darts}. For EngineNAS, we keep the zero operation as convention, while we remove the zero operation for EngineNAS-DST because it inherently has the ability to learn topology in an explicit manner. 

\begin{table}[h]
	\renewcommand\arraystretch{1.2}
	\small
	\begin{center}
		\begin{tabular}{l|c|c}
			\hline
			\multirow{2}*{Operation} & Standard  & Extended  \\
			~ & Space & Space \\
			\hline
			zero & \ding{51} & \ding{51} \\
			\hline
			$3\times3$ separable convolution & \ding{51} & \ding{51}\\
			$5\times5$ separable convolution & \ding{51} & \ding{51}\\
			$3\times3$ dilated separable convolution & \ding{51} & \ding{51}\\
			$5\times5$ dilated separable convolution & \ding{51} & \ding{51}\\
			$3\times3$ max pooling & \ding{51} & \ding{51}\\
			$3\times3$ average pooling & \ding{51} & \ding{51}\\
			identity & \ding{51} & \ding{51}\\
			\hline
			$1\times1$ convolution & - & \ding{51} \\
			$3\times3$ convolution & - & \ding{51} \\
			$1\times3$ then $3\times1$ convolution & - & \ding{51} \\
			$1\times5$ then $5\times1$ convolution & - & \ding{51} \\
			$1\times7$ then $7\times1$ convolution & - & \ding{51} \\
			\hline
		\end{tabular}
	\end{center}
	\caption{The standard and extended operation space.}
	\label{operation}
	\vspace{-4mm}
\end{table}

\section*{Appendix B: Visualization of Searched Architectures}
\label{app_c}
\begin{figure*}
	\centering
	\includegraphics[width=1\linewidth]{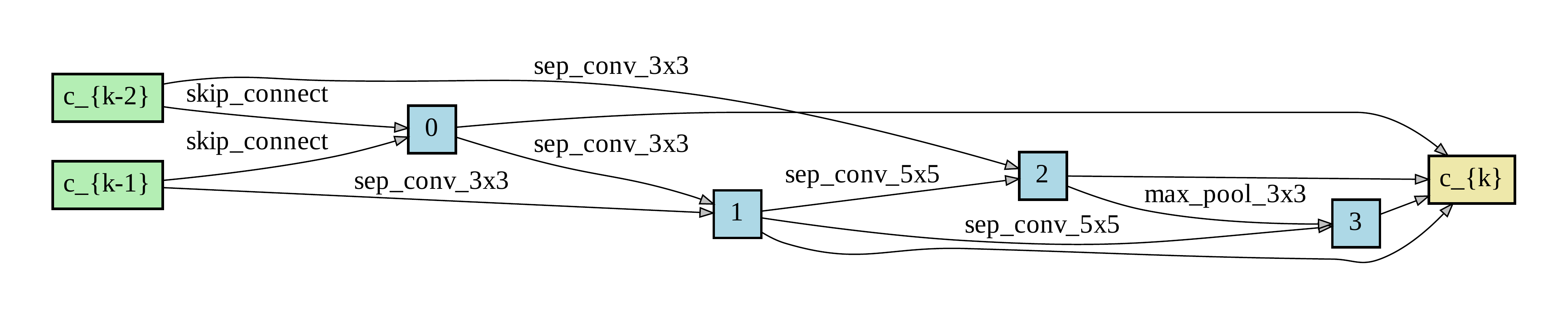}
	\vspace{-7mm}
	\caption{EnTranNAS normal cell searched on CIFAR-10 (the result in Table~\ref{tab1}).}
	\label{c10_normal}
\end{figure*}

\begin{figure*}
	\centering
	\includegraphics[width=0.95\linewidth]{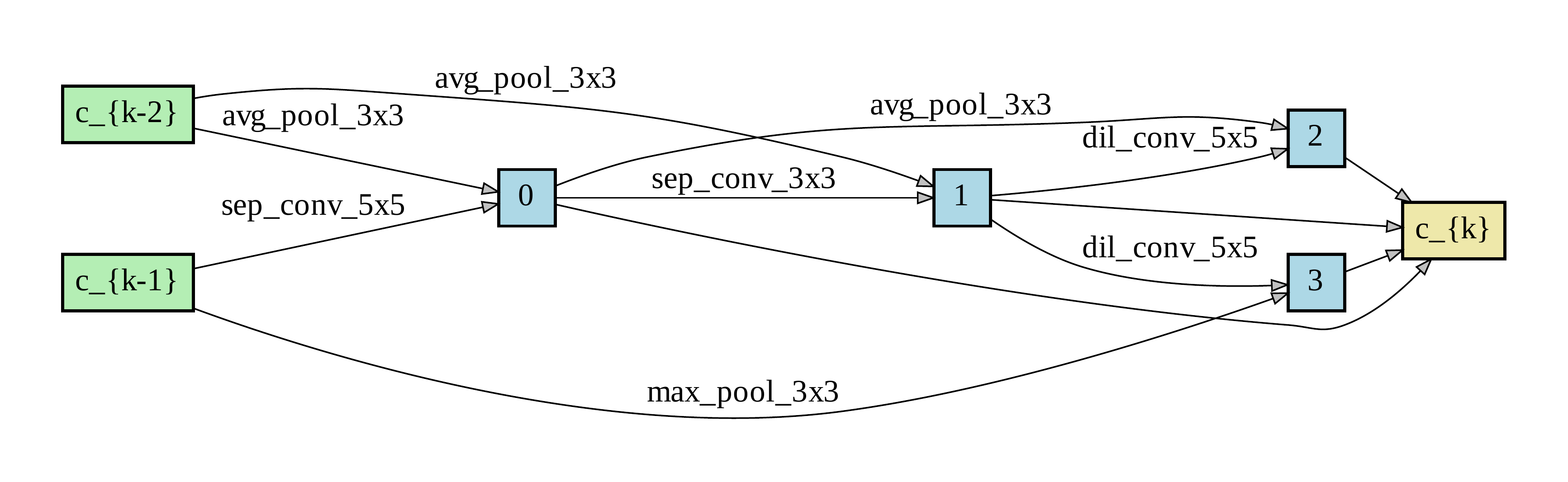}
	\vspace{-5mm}
	\caption{EnTranNAS reduction cell searched on CIFAR-10 (the result in Table~\ref{tab1}).}
	\label{c10_reduce}
\end{figure*}

\begin{figure*}[!t]
	\centering
	\includegraphics[width=0.95\linewidth]{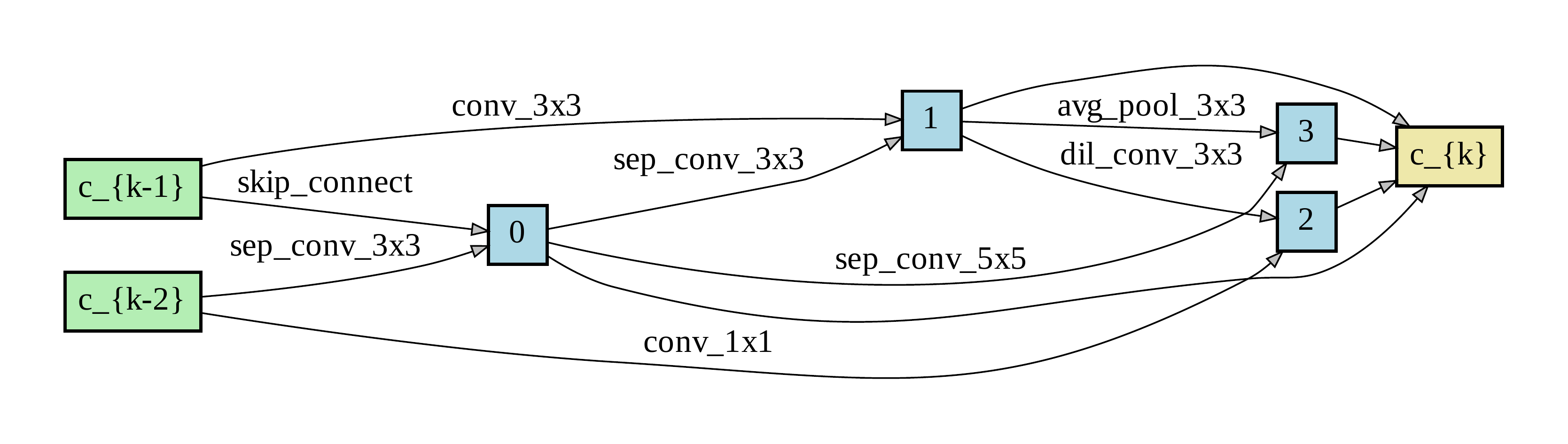}
	\vspace{-5mm}
	\caption{EnTranNAS (12 operations) normal cell searched on CIFAR-10 (the result in Table~\ref{tab1}).}
	\label{c10_12_normal}
\end{figure*}

\begin{figure*}[!t]
	\centering
	\includegraphics[width=0.73\linewidth]{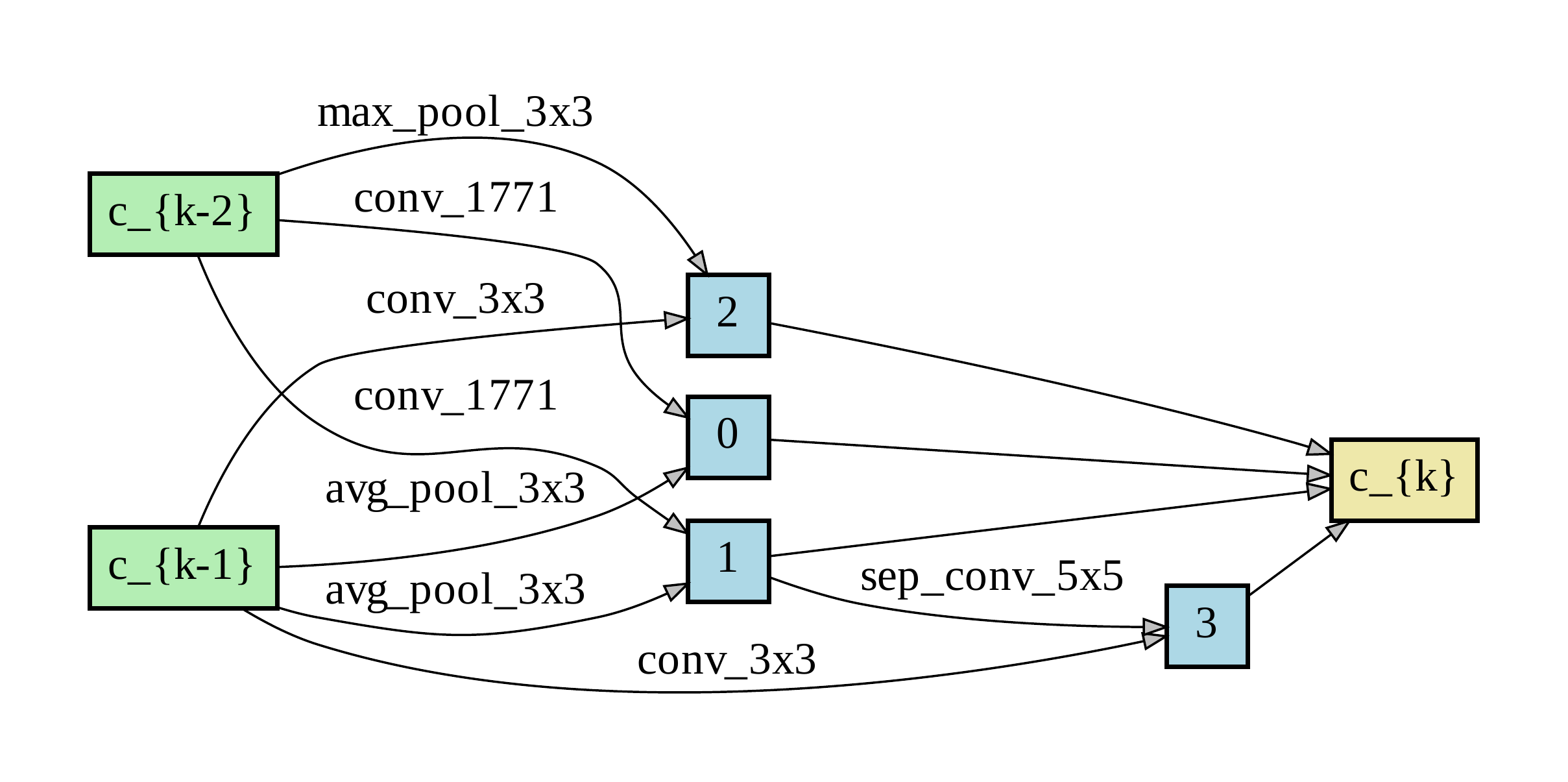}
	\vspace{-5mm}
	\caption{EnTranNAS (12 operations) reduction cell searched on CIFAR-10 (the result in Table~\ref{tab1}).}
	\label{c10_12_reduce}
\end{figure*}

\begin{figure*}
	\centering
	\includegraphics[width=0.7\linewidth]{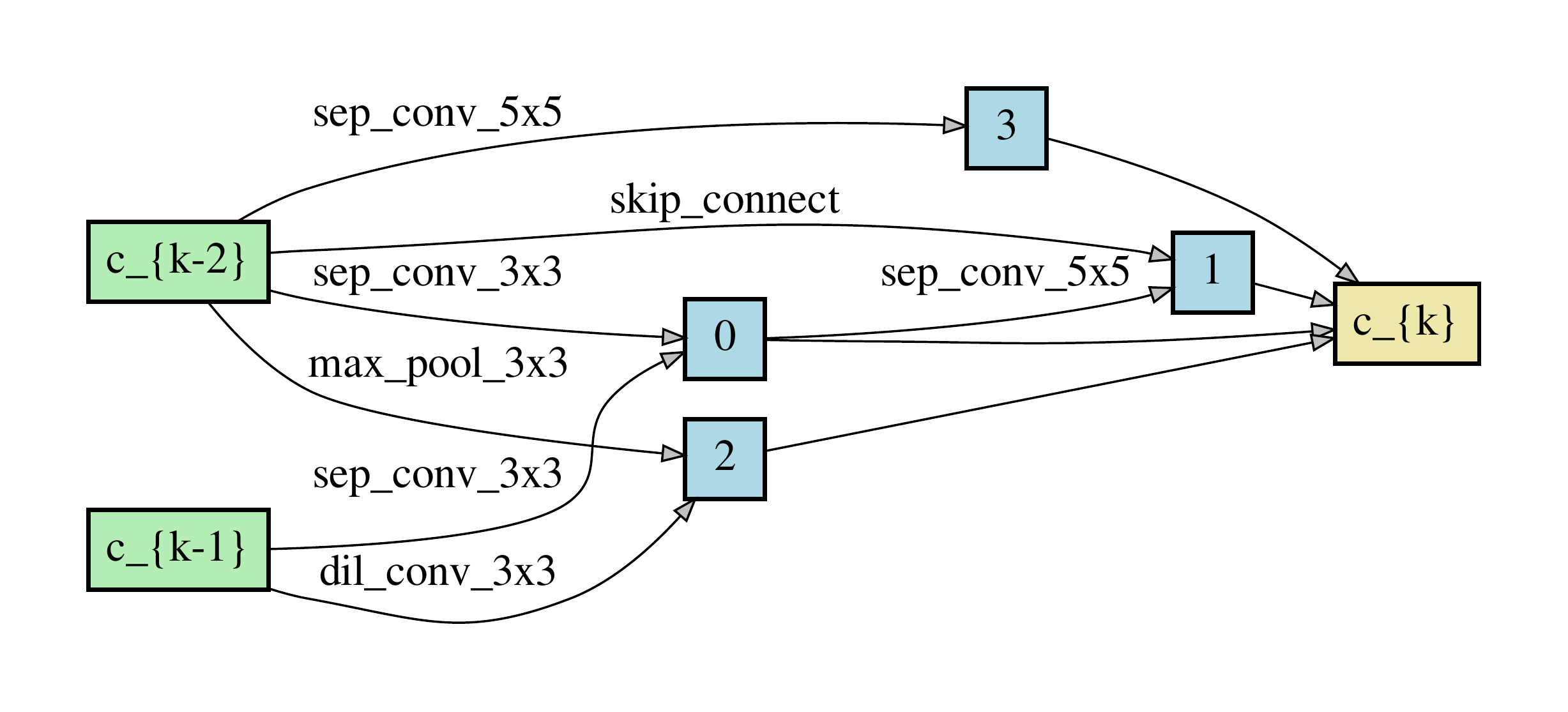}
	\vspace{-4mm}
	\caption{EnTranNAS-DST normal cell searched on CIFAR-10 (the result in Table~\ref{tab1}).}
	\label{c10_dst_normal}
\end{figure*}

\begin{figure*}
	\centering
	\includegraphics[width=0.7\linewidth]{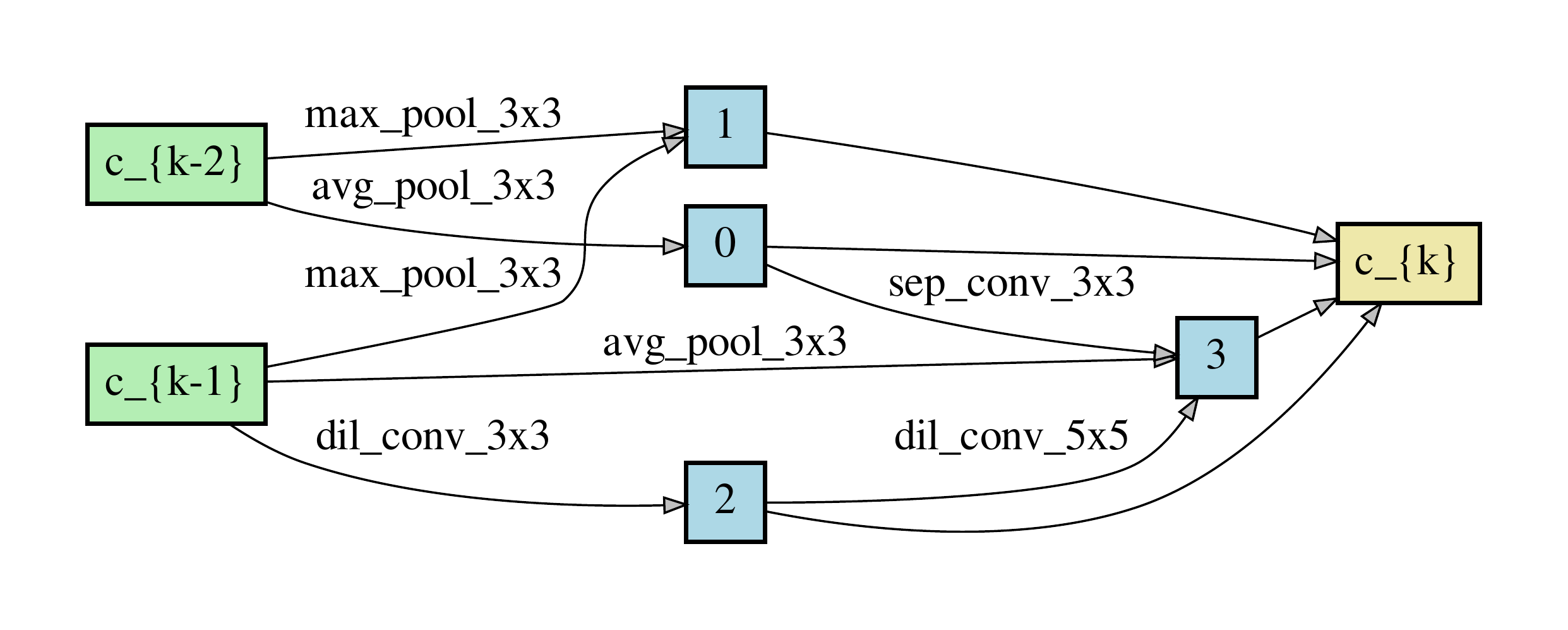}
	\vspace{-4mm}
	\caption{EnTranNAS-DST reduction cell searched on CIFAR-10 (the result in Table~\ref{tab1}).}
	\label{c10_dst_reduce}
\end{figure*}

\begin{figure*}[!t]
	\centering
	\includegraphics[width=0.7\linewidth]{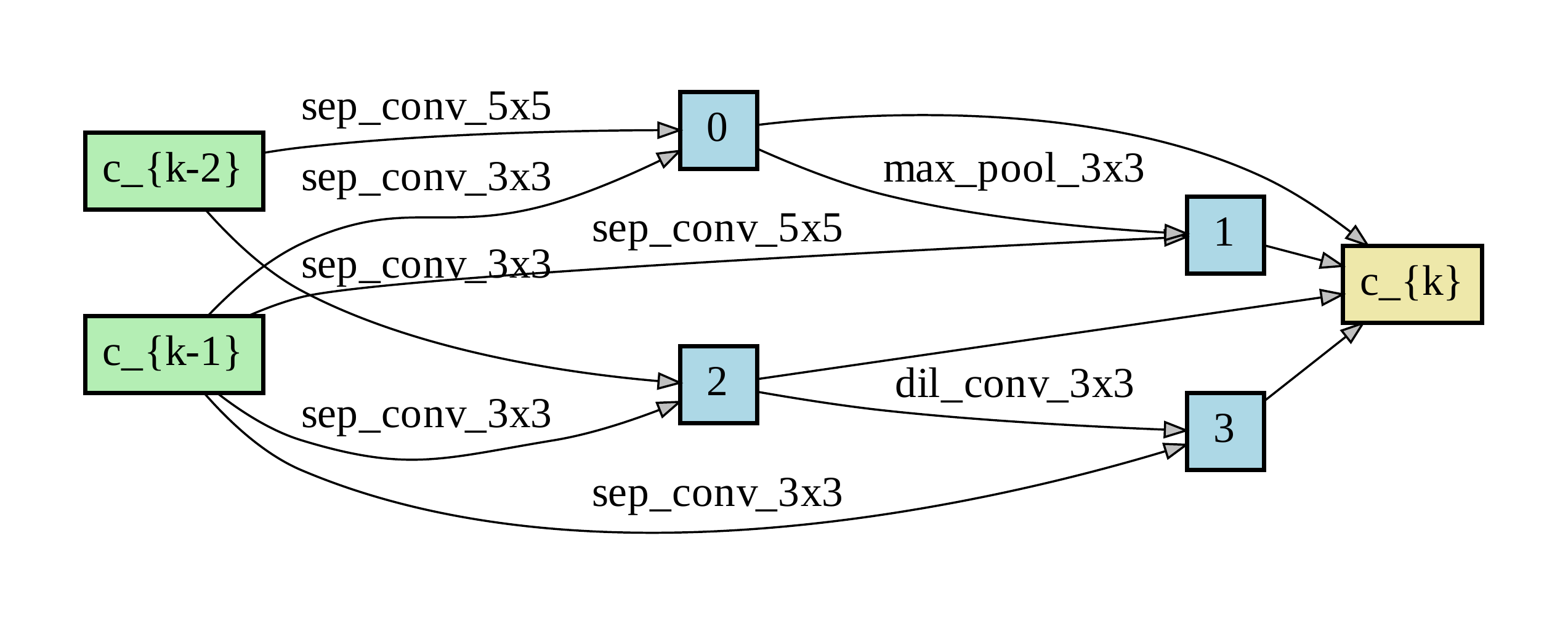}
	\vspace{-4mm}
	\caption{EnTranNAS (ImageNet) normal cell searched on ImageNet (the result in Table~\ref{tab2}).}
	\label{Img_normal}
	\vspace{-3mm}
\end{figure*}

\begin{figure*}[!t]
	\centering
	\includegraphics[width=0.9\linewidth]{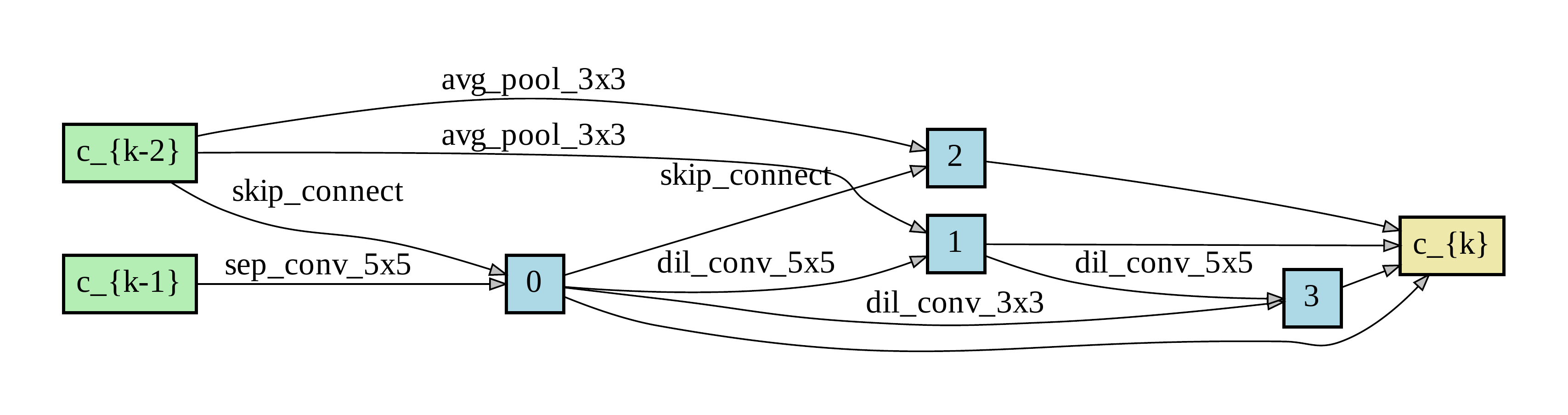}
	\vspace{-4mm}
	\caption{EnTranNAS (ImageNet) reduction cell searched on ImageNet (the result in Table~\ref{tab2}).}
	\label{Img_reduce}
\end{figure*}

\begin{figure*}
	\centering
	\includegraphics[width=0.9\linewidth]{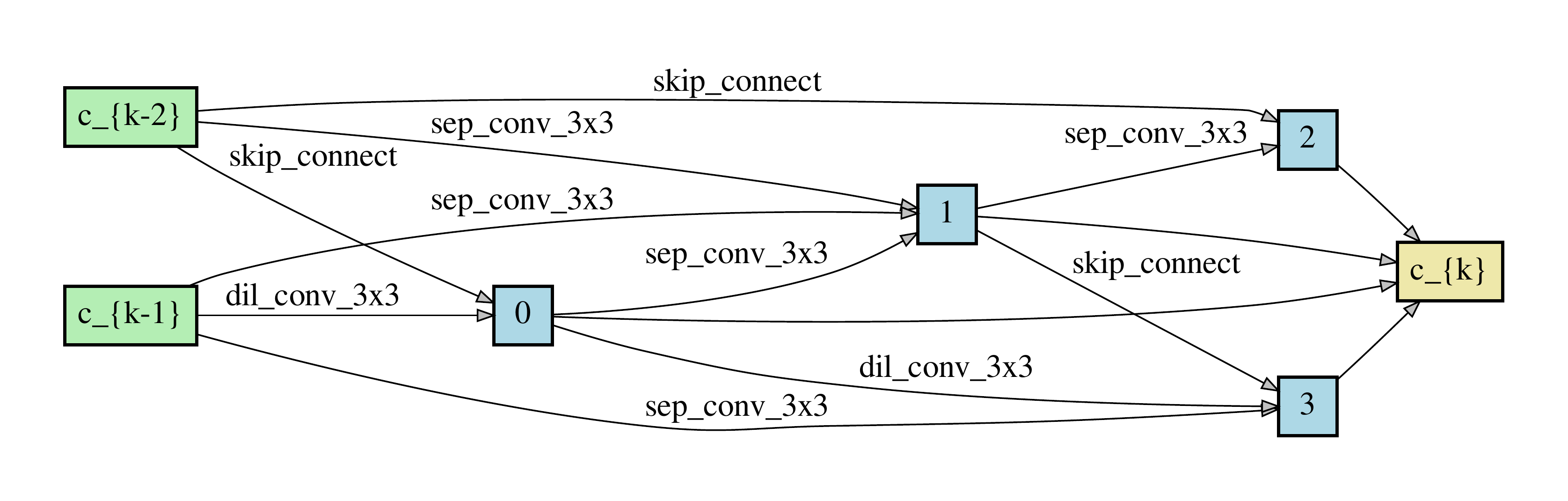}
	\vspace{-4mm}
	\caption{EnTranNAS-DST (ImageNet) normal cell searched on CIFAR-10 (the result in Table~\ref{tab2}).}
	\label{Img_dst_normal}
\end{figure*}

\begin{figure*}
	\centering
	\includegraphics[width=0.9\linewidth]{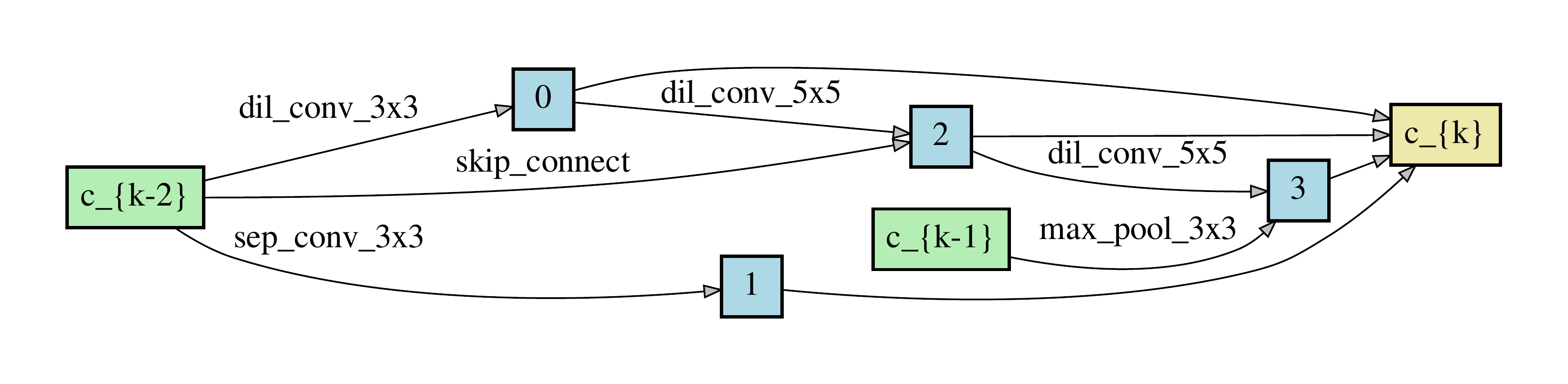}
	\vspace{-4mm}
	\caption{EnTranNAS-DST (ImageNet) reduction cell searched on CIFAR-10 (the result in Table~\ref{tab2}).}
	\label{Img_dst_reduce}
\end{figure*}

\begin{figure*}[!t]
	\centering
	\includegraphics[width=1\linewidth]{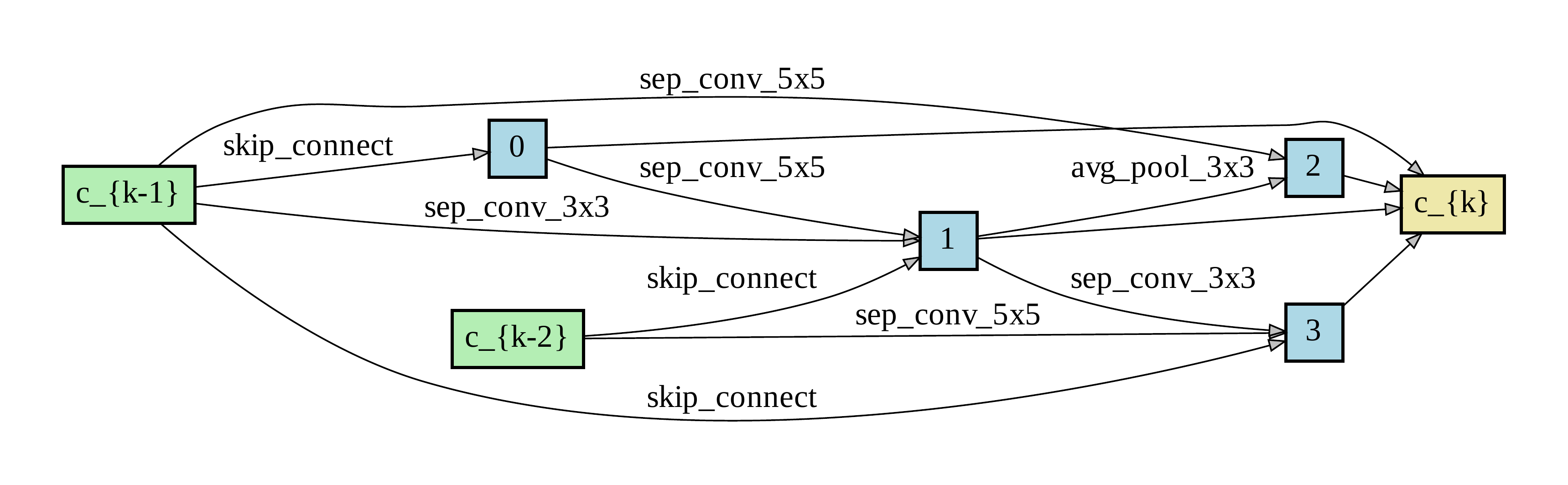}
	\vspace{-8mm}
	\caption{EnTranNAS-DST ($\lambda=0.2$) normal cell searched on ImageNet (the result in Table~\ref{flex}).}
	\label{DST_l02_Img_normal}
\end{figure*}

\begin{figure*}[!t]
	\centering
	\includegraphics[width=0.8\linewidth]{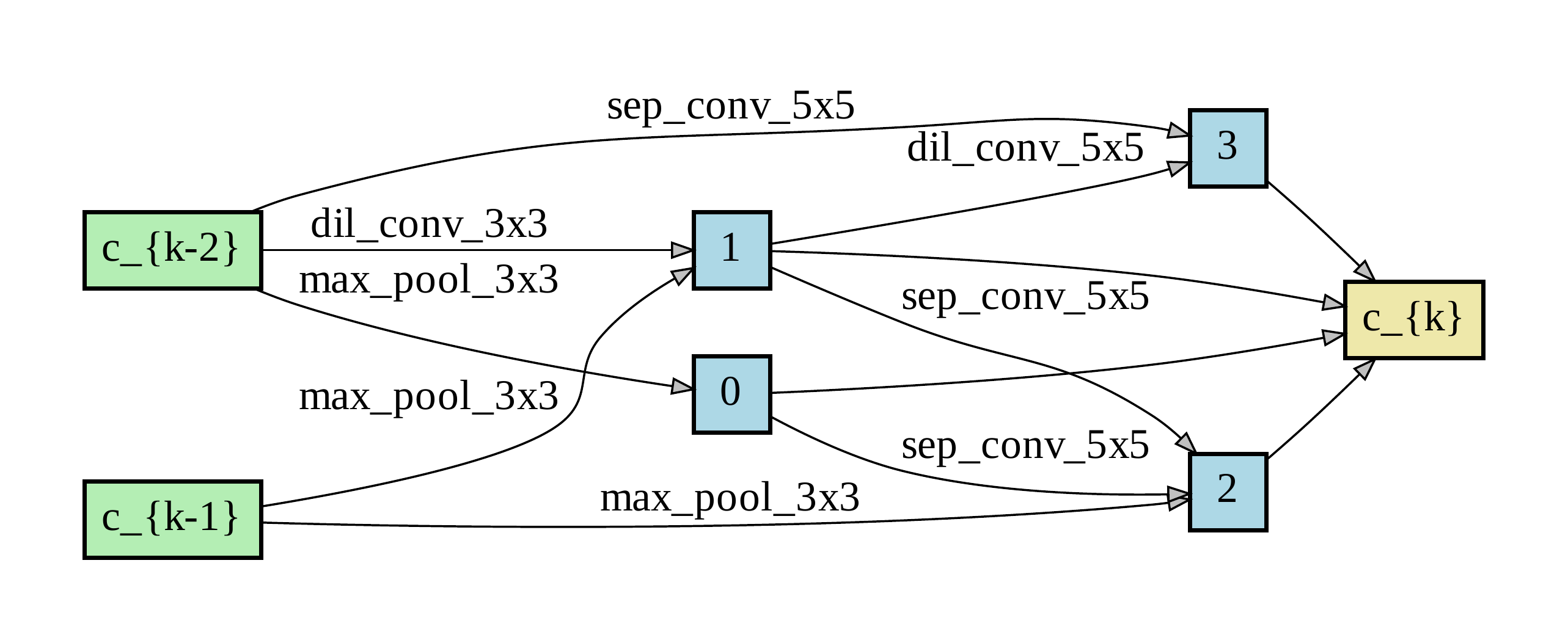}
	\vspace{-4mm}
	\caption{EnTranNAS-DST ($\lambda=0.2$) reduction cell searched on ImageNet (the result in Table~\ref{flex}).}
	\label{DST_l02_Img_reduce}
\end{figure*}

\begin{figure*}[!t]
	\centering
	\includegraphics[width=0.95\linewidth]{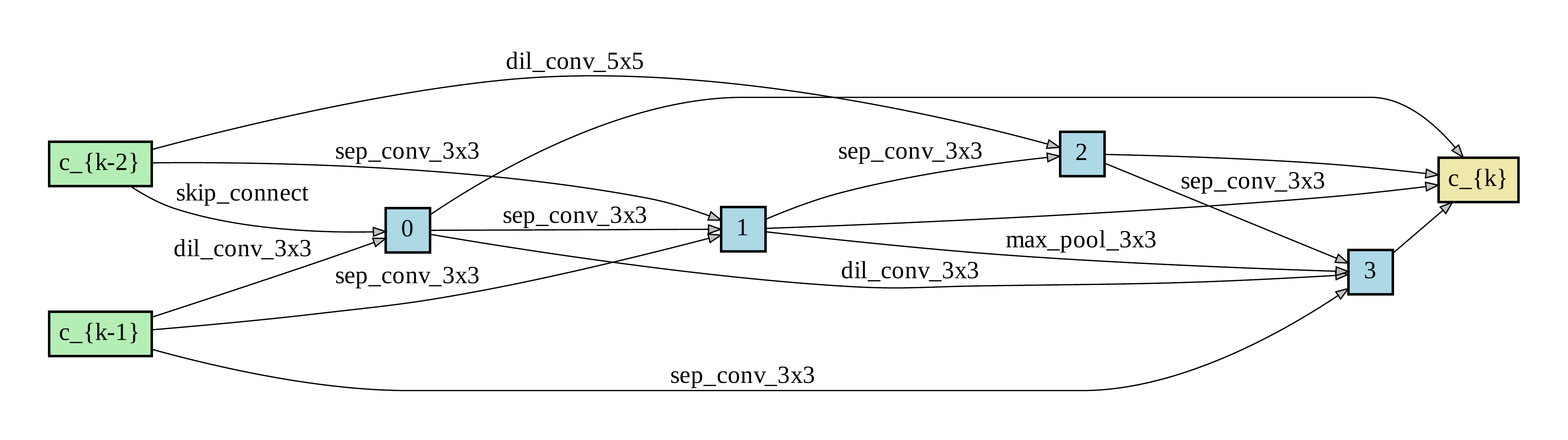}
	\vspace{-4mm}
	\caption{EnTranNAS-DST ($\lambda=0.1$) normal cell searched on ImageNet (the result in Table~\ref{flex}).}
	\label{DST_l01_Img_normal}
\end{figure*}

\begin{figure*}[!t]
	\centering
	\includegraphics[width=0.9\linewidth]{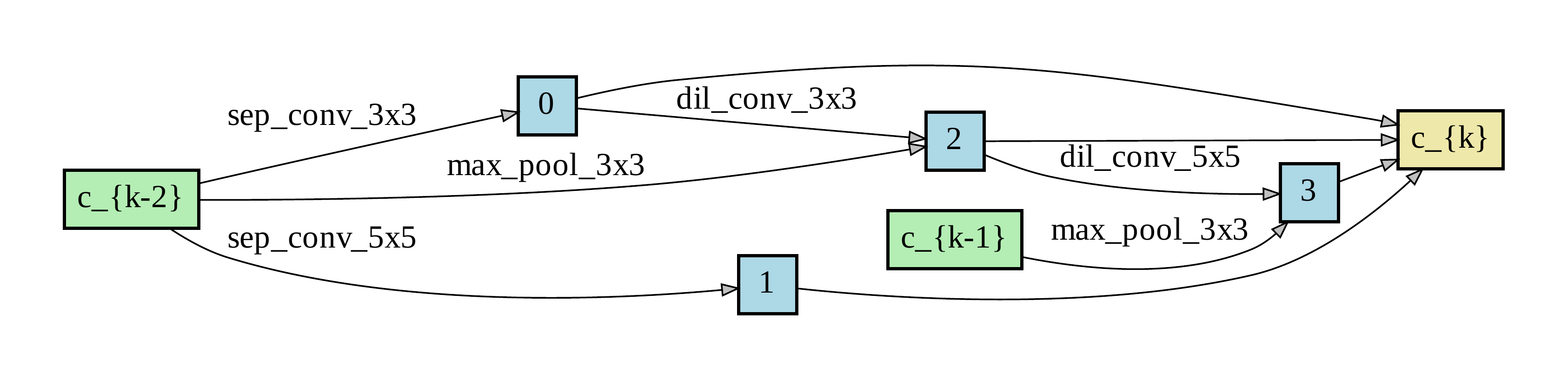}
	\vspace{-4mm}
	\caption{EnTranNAS-DST ($\lambda=0.1$) reduction cell searched on ImageNet (the result in Table~\ref{flex}).}
	\label{DST_l01_Img_reduce}
\end{figure*}

\begin{figure*}[!t]
	\centering
	\includegraphics[width=0.9\linewidth]{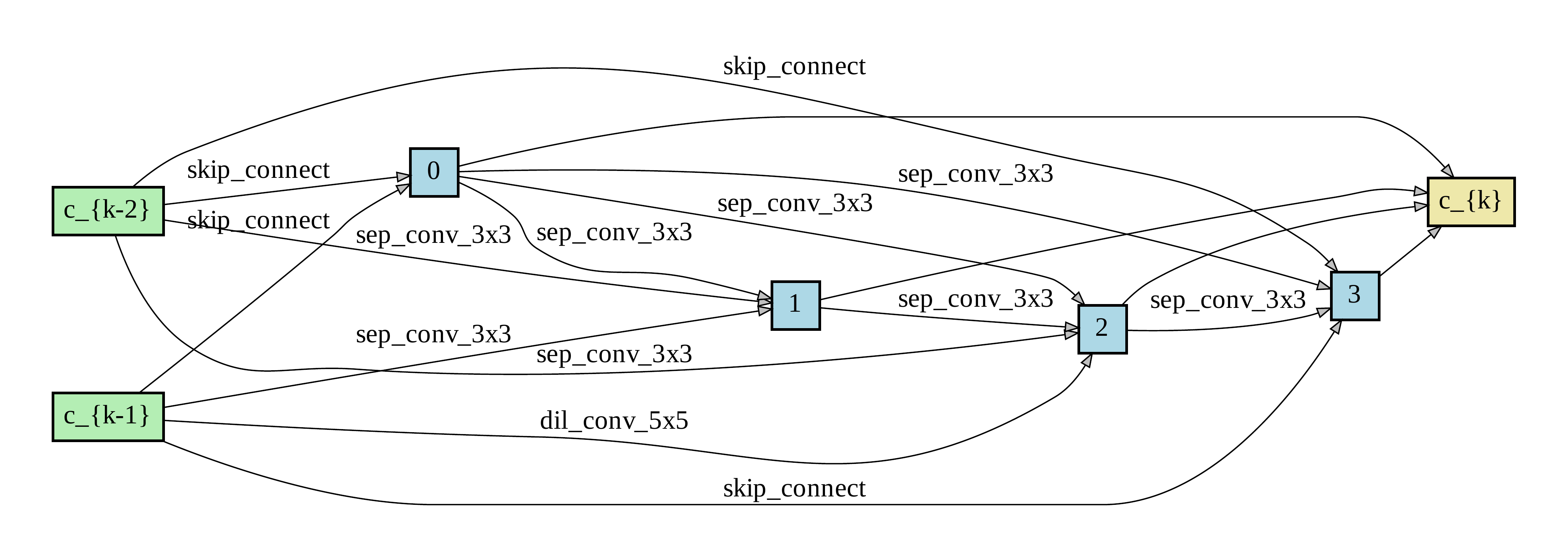}
	\vspace{-4mm}
	\caption{EnTranNAS-DST ($\lambda=0.05$) normal cell searched on ImageNet (the result in Table~\ref{flex}).}
	\label{DST_l005_Img_normal}
\end{figure*}

\begin{figure*}[!t]
	\centering
	\includegraphics[width=0.9\linewidth]{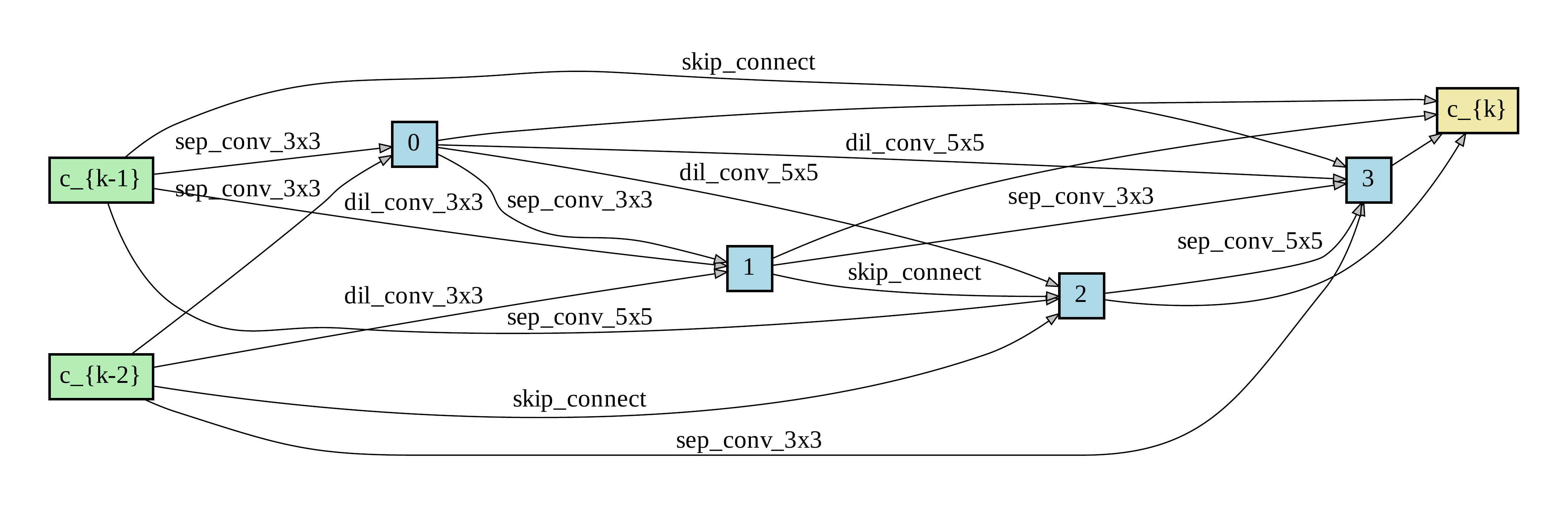}
	\vspace{-4mm}
	\caption{EnTranNAS-DST ($\lambda=0.05$) reduction cell searched on ImageNet (the result in Table~\ref{flex}).}
	\label{DST_l005_Img_reduce}
\end{figure*}

We visualize all searched architectures of our methods. Concretely, the EnTranNAS searched on the standard space of CIFAR-10 is shown in Figure \ref{c10_normal} and \ref{c10_reduce}. Its result on the extended space is shown in Figure \ref{c10_12_normal} and \ref{c10_12_reduce}. The EnTranNAS-DST searched on CIFAR-10 is shown in Figure~\ref{c10_dst_normal} and \ref{c10_dst_reduce}. The EnTranNAS (ImageNet) directly searched on ImageNet is shown in Figure \ref{Img_normal} and \ref{Img_reduce}. The EnTranNAS-DST (ImageNet) on ImageNet is shown in Figure \ref{Img_dst_normal} and \ref{Img_dst_reduce}. In our ablation experiments, the results of EnTranNAS-DST with different $\lambda$ on ImageNet are shown from Figure \ref{DST_l02_Img_normal} to \ref{DST_l005_Img_reduce}.

To better inspect the search process of EnTranNAS-DST, we record the derived architecture of each epoch in the video attached in the supplementary file. It is shown that at the beginning of search, all connections are kept active. That is because by the initialization of $\{\alpha^{(i,j)}_k\}$, $\{p^{(i,j)}_k\}$ have similar values $\forall i<j,1\le k\le K$, and thus $\{\hat{p}^{(i,j)}_k\}$ are large by Eq. (7) and will not be thresholded by Eq. (8), \emph{i.e.}, $\hat{q}^{(i,j)}_k>0,\forall i,j,k$. As the optimization proceeds, connections begin to be cut out gradually. we finally obtain an architecture with both operation and topology learnable. The topology is not subject to any hand-crafted rule.


\newpage
\
\
\
\newpage
\
\
\
\newpage
\ 
\
\

\end{document}